\pdfoutput=1
\documentclass[11pt]{article}
\usepackage{hyperref}
\usepackage{naacl2021}

\usepackage{times}
\usepackage{latexsym}
\usepackage[inline]{enumitem}
\usepackage[T1]{fontenc}
\usepackage[utf8]{inputenc}
\usepackage{enumitem}
\usepackage{microtype}
\def\textttA#1{\texttt{\textls*[-20]{#1}}}

\usepackage{booktabs}
\usepackage[leftmargin=3em]{quoting}
\title{Preregistering NLP research}

\author{Emiel van Miltenburg \and Chris van der Lee \and Emiel Krahmer\\
  Tilburg Center for Cognition and Communnication (TiCC)\\
  Tilburg University\\
  Tilburg, The Netherlands\\
  \textttA{\{C.W.J.vanMiltenburg,C.vdrLee,E.J.Krahmer\}@tilburguniversity.edu}}
\begin{document}
\maketitle
\begin{abstract}
Preregistration refers to the practice of specifying what you are going to do, and what you expect to find in your study, before carrying out the study. This practice is increasingly common in medicine and psychology, but is rarely discussed in NLP. This paper discusses preregistration in more detail, explores how NLP researchers could preregister their work, and presents several preregistration questions for different kinds of studies. Finally, we argue in favour of \emph{registered reports}, which could provide firmer grounds for \emph{slow science} in NLP research. The goal of this paper is to elicit a discussion in the NLP community, which we hope to synthesise into a general NLP preregistration form in future research.
\end{abstract}

\section{Introduction}
Scientific results are only as reliable as the methods that we use to obtain those results. Recent years have seen growing concerns about the reproducibility of scientific research, leading some to speak of a `reproducibility crisis' (see \citealt{sep-scientific-reproducibility} for an overview of the debate). Although the main focus of the debate has been on psychology (e.g.\ through \citealt{aac4716}) and medicine \cite{macleod2014biomedical}, there are worries about the reproducibility of Natural Language Processing (NLP) research as well \cite{fokkens-etal-2013-offspring,cohen-etal-2018-three,moore-rayson-2018-bringing,branco-etal-2020-shared}. The reproducibility debate has led to \citeauthor{munafo2017manifesto}'s (\citeyear{munafo2017manifesto}) \textit{Manifesto for reproducible science}, where the authors discuss the different threats to reproducible science, and different ways to address these threats. We will first highlight some of their proposals, and discuss their adoption rate in NLP. Our main observation is that preregistration is rarely used. We believe this is an undesirable situation, and devote the rest of this paper to argue for preregistration of NLP research.

\citeauthor{munafo2017manifesto} recommend \textbf{more methodological training}, so that e.g.\ statistical methods are applied correctly. In NLP, we see different researchers picking up the gauntlet to teach others about statistics \cite{dror-etal-2018-hitchhikers,dror2020statsbook}, achieving language-independence \cite{bender2011achieving}, or best practices in human evaluation \cite{van-der-lee-etal-2019-best,VANDERLEE2021101151}. Moreover, every *ACL conference offers tutorials on a wide range of different topics. While efforts to improve methodology could be more systematic (e.g.\ by actively encouraging methodology tutorials, and working towards community standards),\footnote{A more radical proposal would be to \emph{always} host methodology-focused tutorials, and to invite researchers to teach specific modules, similar to keynote talks.} the infrastructure is in place.

\citeauthor{munafo2017manifesto}\ also recommend to \textbf{diversify peer review}. Instead of only having journals, that are responsible for both the evaluation and dissemination of research, we can now also solicit peer feedback after publishing our work on a platform like ArXiv or OpenReview. The NLP community is clearly ahead of the curve in terms of the adoption of preprints, and actively discussing ways to improve peer review (\citealt{longterm,shortterm,rogers2020peer}). To improve the quality of the reviews themselves, ACL2020 featured a tutorial on peer reviewing \cite{cohen-etal-2020-reviewing}.

Another advice from \citeauthor{munafo2017manifesto} is to \textbf{adopt reporting guidelines}, so that papers include all relevant details for others to reproduce the results. The NLP community is rapidly adopting such guidelines, in the form of \citeauthor{dodge-etal-2019-show}'s (\citeyear{dodge-etal-2019-show}) reproducibility checklist that authors for EMNLP2020 need to fill in. Beyond reproducibility, we are also seeing more and more researchers adopting Data statements \cite{bender-friedman-2018-data}, Model cards \cite{mitchell.etal2019}, and Datasheets \cite{gebru.etal2018} for ethical reasons.

\begin{table*}
\centering\small
    \begin{tabular}{lp{11.75cm}}
\toprule
\textbf{Data collection} & Have any data been collected for this study already?\\
\textbf{Hypothesis} & What's the main question being asked or hypothesis being tested in this study?\\
\textbf{Dependent variable} & Describe the key dependent variable(s) specifying how they will be measured.\\
\textbf{Conditions} & How many and which conditions will participants be assigned to?\\
\textbf{Analyses} & Specify exactly which analyses you will conduct to examine the main question/hypothesis.\\
\textbf{Outliers and Exclusions} & Describe exactly how outliers will be defined and handled, and your precise rule(s) for excluding observations.\\
\textbf{Sample Size} & How many observations will be collected or what will determine sample size?\\
\textbf{Other} & Anything else you would like to pre-register?\\
\midrule
\textbf{Research aim} & Specify the overall aim of the research.\\
\textbf{Use of literature} & Specify the role of theory in your research design.\\
\textbf{Rationale} & Elaborate if your research is conducted from a certain theoretical perspective.\\
\textbf{Tradition} & Specify the type of tradition you work in: grounded theory, phenomenology, \ldots\\
\textbf{Data collection plan} & Describe your data collection plan freely. Be as explicit as possible.\\
\textbf{Type of data collected} & Select the type(s) of data you will collect.\\
\textbf{Type of sampling} & Indicate the type of sampling you will rely on: purposive, theoretical, convenience, snowball\ldots\\
\textbf{Rationale} & Indicate why you choose this particular type of sampling.\\
\textbf{Sort of sample} & Pick the ideal composition of your sample: heterogenous, homogenous, \ldots\\
\textbf{Stopping rule} & Indicate what will determine to stop data collection: saturation, planning, resources, other.\\
\textbf{Data collection script} & Upload your topic guide, observation script, focus group script, etc.\\
\bottomrule
    \end{tabular}
    \caption{Top: preregistration form from AsPredicted (\url{https://aspredicted.org}) for quantitative research. Bottom: additional items from \citet{doi:10.1080/08989621.2019.1580147} for qualitative research. All text is quoted verbatim.}
    \label{tab:aspredicted}
\end{table*}

\citeauthor{munafo2017manifesto}'s final recommendation, \textbf{preregistration}, means that authors should specify what they are going to do, and what they expect to find, before carrying out their studies \cite{Nosek2600}. The goal of preregistration is to ensure that all hypotheses and research methods are made explicit before researchers are confronted with the data. Otherwise, researchers end up in a \textit{garden of forking paths}, where all research decisions are made implicitly, based on common sense and the available data \cite{gelman2013garden}. This negatively impacts the reliability and generalisability of any study. In other words: preregistration allows us to distinguish between exploratory and confirmatory research. Exploratory research does not require preregistration, because the goal is to get a sense of what is possible. Any pattern you come across during exploratory research, allows us to draw up hypotheses. For a subsequent confirmatory study you could/should preregister to test those hypotheses. By explicitly marking (parts of) your study as exploratory or confirmatory, it is easier to understand the status of your results.

Compared to the work on reporting quality, there has been little talk of preregistration in the NLP literature; the terms `preregister' or `preregistration' are hardly used in the ACL Anthology.\footnote{Looking for these terms, we found four papers that mention preregistration: \citet{cao-etal-2018-babycloud} and \citet{van-der-lee-etal-2019-best} mention it, and \citet{van-miltenburg-etal-2018-didec} and \citet{futrell-levy-2019-rnns} share their own preregistration.} For this reason, we will focus on preregistration and its application in NLP research. The next sections discuss how preregistration works (\S\ref{sec:how}), propose preregistration questions for NLP research (\S\ref{sec:form}), discuss the idea of `registered reports' as an alternative pathway to publication (\S\ref{sec:regreports}) and the overall feasibility of preregistrations in NLP (\S\ref{sec:feasibility}).

\section{How does preregistration work?}
\label{sec:how}
%\urlstyle{rm}
Before you begin, you enter the hypotheses, design, and analysis plan of your study on a website like  \href{https://osf.io/}{the Open Science Framework}, \href{https://aspredicted.org}{AsPredicted}, or \href{https://researchbox.org}{ResearchBox}. These sites provide a time stamp; evidence that you indeed made all the relevant decisions before carrying out the study. During your study, you follow the preregistered plans as closely as possible. In an ideal world, there would be an exact match between your plans and the actual study you carried out. But there are usually unforeseen circumstances that force you to change your study. This is fine, if the changes are clearly specified (including the reasons for those changes) in your final report \cite{Nosek2600}.

\textbf{A typical preregistration form.}
Table~\ref{tab:aspredicted} shows questions from the preregistration form from AsPredicted.\footnote{See \url{https://osf.io/zab38/wiki/home/} for an overview of different forms.} This form is geared towards hypothesis-driven, experimental research where human participants are assigned to different experimental conditions. \citet{simmons2017properly} note that answers should state exactly how the study will be executed, but also that it should be short and easy to read.

\begin{table}
    \centering\small
    \begin{tabular}{p{0.94\linewidth}}
    \toprule
    What are your hypotheses/key assumptions?\\
    What is the independent variable? (e.g.\ model architecture)\\
    What is the dependent variable (e.g.\ output quality)\\
    How will you measure the dependent variable?\\
    Is there just one condition (corpus/task), or more?\\
    What parameter settings will you use?\\
    What data will you use, and how is it split in train/val/test?\\
    Why this data? What are key properties of the data?\\
    How will you analyse the results and test the hypotheses?\\
    \bottomrule
    \end{tabular}
    \caption{Questions for analysis, experiments, and reproduction papers (expanded in Appendix~\ref{sec:forms}).}
    \label{tab:nlpquestions}
\end{table}

\textit{Data collection, hypothesis, dependent variable.} The form first asks whether data collection has been carried out yet (ideally the answer should be \textit{no}, but see Appendix \S\ref{sec:availabledata}), and then asks researchers to make their main hypothesis explicit so that it cannot be changed after the fact. Following the hypothesis, researchers should describe their key dependent variables (i.e.\ the main outcome variables) and how they will be measured. This includes cutoff points that will be used to discretise continuous variables (e.g.\ to divide participants in different groups). 

\textit{Conditions, analyses, outliers and exclusions.} Next, the form asks about the design of the study, the analyses, and the process of determining outliers (and whether those should be excluded). The answer needs to be detailed enough so that other researchers are able to reproduce the study.

\textit{Sample size \emph{and} other.} The form then asks how much data will be collected, so as to prevent \textit{optional stopping} (where researchers keep collecting data until the results are in line with their preferred hypothesis).\footnote{Although it is not necessary for the form, at this point it is good to justify the sample size, e.g. by using a power analysis.} Finally, the form allows researchers to specify other aspects of the study they would like to preregister, such as ``secondary analyses, variables collected for exploratory purposes, [or] unusual analyses.'' 

\textbf{Qualitative research.}
Preregistration is not only suitable for quantitative research; \citet{doi:10.1080/08989621.2019.1580147} present a proposal to preregister qualitative studies as well. Their suggestions are also presented in Table~\ref{tab:aspredicted}. The authors argue that, although qualitative research differs in its goals from quantitative research (developing theories rather than testing them), it is still valuable to make your assumptions and research plans explicit before carrying out your planned study. Because qualitative research is more flexible than quantitative research, \citeauthor{doi:10.1080/08989621.2019.1580147} view qualitative preregistrations as living documents; continuously updated to track the research progress. This stimulates conscientiousness, and avoids sloppy research. Public preregistrations also allow for immediate feedback.

\section{Preregistration in NLP research}
\label{sec:form}
To determine what a preregistration for NLP research should look like, we need to consider the different kinds of research contributions in NLP. For this, we use the paper types proposed for COLING \citeyear{coling-2018-international}.\footnote{\url{https://coling2018.org/paper-types/}} These are: Computationally-aided linguistic analysis; NLP engineering experiment paper; Reproduction/Resource/Position/Survey Paper.
Of these, position papers are less suitable for preregistration, since these are more opinion/experience-driven, and the process of writing them cannot be formalised. We treat the others below.

\begin{table}
    \centering\small
    \begin{tabular}{p{0.94\linewidth}}
    \toprule
    What do you aim to learn from the error analysis?\\
    What do you know from the literature about system errors?\\
    What kinds of errors do you expect to find?\\
    How will you sample the outputs to analyse?\\
    Do you also consider the input in your sampling strategy?\\
    How do you plan to analyse the output?\\
    How many judges will assess the output? Are they trained?\\
    How is the reliability of the judges assessed?\\
    Is there a fixed error categorisation scheme or not?\\
    \bottomrule
    \end{tabular}
    \caption{Questions to ask before an error analysis.}
    \label{tab:erroranalysis}
\end{table}

\textbf{Analysis, experiments, and reproduction papers} typically have one or more hypotheses, even though they may not always be marked as such.\footnote{Taking the best papers from COLING 2018 as an example, \citet[analysis]{ruppenhofer-etal-2018-distinguishing} test assumptions from the linguistics literature about affixoids, \citet[experiment]{thompson-mimno-2018-authorless} test which subsampling methods improve the output generated by topic models, and \citet[reproduction]{lan-xu-2018-neural} test whether the reported performance for different neural network models generalises to other tasks.} This means we can ask many of the same questions for these studies as for experimental research. Table~\ref{tab:nlpquestions} provides a rough overview of important questions to ask before carrying out your research.

If your study contains an error analysis, then you could ask the more qualitatively oriented questions in Table~\ref{tab:erroranalysis}. They acknowledge that you always enter error analysis with some expectation (i.e.\ researcher bias) of what kinds of mistakes systems are likely to make, and where those mistakes may be found. The questions also stimulate researchers to go beyond the practice of providing some `lemons' alongside cherry-picked examples showing good performance.

The main benefit of asking these questions beforehand is that they force researchers to carefully consider their methodology, and they make researchers' expectations explicit. This also helps to identify unexpected findings, or changes that were made to the research design during the study.

\textbf{Resource papers} are on the qualitative side of the spectrum, and as such the questions from \citet{doi:10.1080/08989621.2019.1580147}, presented at the bottom of Table~\ref{tab:aspredicted}, are generally appropriate for these kinds of papers as well. Particularly 1) the original purpose for collecting the data, 2) sampling decisions (what documents to include), and 3) annotation (what framework/perspective to use) are important. Because the former typically influences the latter two, it is useful to document how the goal of the study influenced decisions regarding sampling and annotation, in case the study at some point pivots towards another goal.

\textbf{Survey papers} should follow the PRISMA guidelines for structured reviews \cite{10.1371/journal.pmed.1000097,10.1371/journal.pmed.1000100}. According to these guidelines, researchers should state exactly where they searched for existing literature, what search terms they used, and what criteria they used to select relevant papers. This increases reproducibility, allows readers to find any gaps in the survey, and avoids a biased presentation of the literature (i.e.\ only citing researchers you know, or work that fits your preferred narrative). A recent NLP example of a structured review is provided by \citet{reiter-2018-structured}.

\section{Registered reports}
\label{sec:regreports}
Registered reports ``[split] conventional peer review in half'' \cite{chambers2019next}. First, authors submit a well-motivated research plan for review, before carrying out the study (similar to a preregistration). This plan may go back-and-forth between the authors and the reviewers, but once the plan is accepted, the authors receive the guarantee that, if they carry out the study according to plan, their work will be published. As with preregistration, deviations from the original plan are allowed, but these should be indentified in the final report. The main advantage of registered reports is that they provide a means to avoid publication bias. Because studies aren't judged on the basis of their results, positive results are equally likely to be published as negative results. As long as the study is deemed valuable \textit{a priori}, it should get published. An additional benefit of registered reports is that reviews may actually correct flaws in the research design, meaning that we reduce the chance of running an expensive study all for nothing. In the case of NLP research, this may save a lot of energy (cf.\ \citealt{strubell-etal-2019-energy}). We are not aware of any NLP journals that offer registered reports, but strongly encourage the NLP community to take steps in this direction.\footnote{Cf.\ \citet{ijcai2018-717} regarding AI research.}

\section{Feasibility}
\label{sec:feasibility}
\citet{gelman2013garden,gelman2014statistical} touch upon the feasibility of preregistration, noting that: 
\begin{quoting}[leftmargin=\parindent]\small
``[f]or most of our own research projects this strategy hardly seems possible: in our many applied research projects, we have learned so much by looking at the data. Our most important hypotheses could never have been formulated ahead of time.''
\end{quoting}

This certainly rings true for NLP as well.
However, we should be careful about conclusions that are drawn on the basis of pre-existing data. \citet{gelman2013garden} note that in such cases, if it is feasible to collect more data, it is good to follow up positive results with a pre-registered replication to confirm your initial findings. One way to do this is to collect and evaluate your model on a new test set (cf.\ \citealt{pmlr-v97-recht19a}). This tells us to what extent trained models generalise to unseen data. Another idea could be to preregister the human evaluation (or error analysis) of the model output.

We believe that preregistration, and especially registered reports, could ease the pressure to publish as soon as possible. If your analysis plan is accepted for publication, you can take as long as you want to actually carry out the study, without having to worry about being scooped. This provides new opportunities for \textit{slow science} in NLP (also see Min-Yen \citeauthor{kan2018research}'s keynote at COLING \citeyear{kan2018research}). 

\section{Questions about preregistration}
Below we address some common questions about preregistration. We thank our anonymous reviewers for raising some of these questions.

\noindent \textbf{Is preregistration more work?} In our experience, preregistration adds little overhead to a research project. Especially if a project requires approval by an Institutional Review Board (IRB), you need to write a description along similar lines anyway. For projects not requiring IRB approval, it is good practice to provide a model card \cite{mitchell.etal2019}, data sheet  \cite{gebru.etal2018} or data statement \cite{bender-friedman-2018-data} with your model or resource. Given the ethical aspects of NLP research, it is advisable to consider all dimensions of your study before you carry it out. Moreover, preregistration is a good way to start writing the paper before carrying out the research, a practice advocated by \citet{eisner2010write} to maximise the impact of your work. Finally, it may be more work to prepare a registered report, but this comes with the benefit of having a pre-approved methodology. Once the project is completed, reviewers will not reject your paper based on methodological choices.

\noindent \textbf{Should I worry about being scooped?}
There is no need to worry. We already discussed registered reports, where research proposals are provisionally accepted before data collection starts. Otherwise, this worry has been addressed through the existence of both public and private preregistrations. A researcher can choose to keep a preregistration private until the research is completed. They can make their preregistration public whenever they like, for example to invite feedback from the community. In addition, preregistrations are also time-stamped, and you can use these time stamps during the review phase to show that you have had these ideas before similar work was published.\footnote{The public/private distinction has been implemented by both the Open Science Foundation and AsPredicted.org. The Open Science Foundation allows for a 4-year embargo, during which the preregistration is kept private. Aspredicted allows for preregistrations to be private indefinitely.}

\noindent \textbf{What about citing preregistrations?}
In some regards, the discussion about preregistrations is similar to the discussion about preprints (i.e.\ papers on ArXiv), thus similar questions arise. Both preregistrations and published studies are being cited. For example, medical journals like BMC Public Health also publish study protocols (similar to preregistrations), without any results, that are also cited by others (e.g.\ work using a similar protocol).

\textbf{What should we do with concurrent work?}
It may of course happen that multiple researchers have similar ideas around the same time. We believe that it is still valuable to publish multiple independent studies with similar results. Even if they don’t provide any new insights (which is rare), they do provide evidence towards the robustness of the findings. Where and how those findings should be published is a separate discussion.\footnote{However, if there is value in publishing the `first' paper, there is probably also value in publishing the `second' one. The same holds for the question of whether both studies should be cited; good scholarship considers \emph{all} the available evidence.}

\noindent \textbf{How should we teach preregistration?}
Preregistration is already being incorporated into Psychology courses (see, for example, \citealt{blincoe2020research}). It is relatively straightforward to implement as part of student research proposals during applied courses in NLP: specify what you plan to do exactly, and what you expect to find. It is often useful for students to have an explicit format to think through their research plans, to make sure that they make sense.

\section{Limitations}
Although preregistration is offered as a solution to  improve our work, it does not solve all of our problems. \Citet{VANTVEER20162} mention three limitations: \textit{1.~Flexibility.} It may be difficult or infeasible for authors to foresee all possible outcomes, and as such there may be gaps in the preregistration, which still allow for flexibility in the analysis.
\textit{2.~Fraud.} There is no way to prevent fraudulent researchers from, e.g., creating multiple preregistrations, or falsely `preregistering' studies that were already run. At some point we just have to trust each other to do the right thing, but increased transparency does make it harder to commit fraud.
\textit{3.~Applicability.} Preregistration may not be possible for all kinds of studies. As discussed above, it has mainly been developed for quantitative studies (particularly experiments), and there are proposals for the preregistration of qualitative research \cite{doi:10.1080/08989621.2019.1580147}, although we have yet to see whether this idea will catch on. Finally, \citet{szollosi2020preregistration} argue that, although preregistration might offer greater transparency, it does not by itself improve scientific reasoning and theory development. Since large parts of NLP are pre-theoretical (we have observed effects but do not have any theoretical explanations for why these effects occur), one might reasonably argue that we should focus on theory development first, before we can carry out any meaningful experiments.%\footnote{Cf. \citeauthor{rahimi2017reflections}'s (\citeyear{rahimi2017reflections}) infamous talk, comparing the state of modern Machine Learning to Alchemy, i.e.\ the times before the development of Chemistry.}

\section{Conclusion}
We have discussed how preregistration could benefit NLP research, and how different kinds of contributions could be preregistered. We have also proposed an initial list of questions to ask before carrying out NLP research (and see Appendix~\ref{sec:forms} for example preregistration forms). With this paper, we hope to encourage other NLP researchers to consider preregistering their work, so that they will no longer get lost in the garden of forking paths. Still, there is no silver bullet to cure sloppy science. Although preregistration is certainly helpful, it does not guarantee high-quality research, and we do need to stay critical about preregistered studies, and the way they are carried out.

\section*{Acknowledgments}
Thanks to the anonymous reviewers for their constructive feedback, and to all the \texttt{\#NLProc} Twitter people for discussion.

\bibliography{anthology,custom}
\bibliographystyle{acl_natbib}

\appendix

\section{Preregistration forms}\label{sec:forms}
This appendix provides preregistration forms for different kinds of paper types. These forms are preliminary, and they are mainly meant as a starting point for discussions of preregistration in NLP. We are happy to admit that there may be flaws in this appendix (either in the forms or in our reasoning). Future work should investigate whether these forms are complete (i.e.\ limit \emph{researcher degrees of freedom} as much as possible) and appropriate for different kinds of NLP research.

\subsection{Preface: data availability in NLP}\label{sec:availabledata}
Preregistration is a means to avoid hindsight bias, because you have to specify your expectations upfront, when your perspective is not yet colored by your experience with the data. But for NLP studies it is unclear what `the data' is. We can distinguish three kinds of data: 1.\ The training/validation/test sets, 2.\ The model output, 3.\ Human judgments.

In an ideal situation, preregistration would occur before any kind of data has been obtained. The problem is that this is often not the case; there are many canonical datasets for which the data is publicly available. Of course one could collect an additional test set (as we suggested above), but the community often judges new approaches based on their performance for established datasets. So what should we do? Still preregister! Arguably the training, validation, and test data is usually not central to the work. What matters is how a particular system performs. So even if we don't usually find ourselves in the ideal situation where none of the data is available yet, it is typically fine to preregister your study if the train/eval/test data is available but system outputs and evaluation scores are not. When authors are transparent in their data sharing policy, we can reconstruct the timeline of events before and after the preregistration, to see how much their knowledge about the data may have influenced them.

\subsection{Computationally aided linguistic analysis}
This paper type corresponds to several different setups, ranging from experiments with human subjects, to corpus analyses to see if particular generalisations from the literature hold up. Preregistration has been discussed from a linguistics perspective by \citet{roettger2020preregistration}. For experiments with human participants, readers may refer to the standard preregistration forms from \href{https://aspredicted.org}{AsPredicted} (see our Table~\ref{tab:aspredicted}), \href{https://osf.io}{OSF}, or the questions from \citeauthor{roettger2020preregistration}'s Figure~1.

For more corpus-oriented studies (e.g.~\citealt{ruppenhofer-etal-2018-distinguishing}), we should consider a mix of the quantitative and qualitative questions from our Table~\ref{tab:aspredicted}. Usually these kinds of studies do require some data collection, so authors should ask:

\begin{enumerate}[noitemsep,topsep=0pt]
    \item What is the goal of this study?
    \item What are the main questions/hypotheses?
    \item What kind of data will be collected?
    \item How will this data be collected?
    \item What sampling strategy will be used? Why?
    \item How much data are you planning to collect?
    (Is there any target or stopping criterion?)
    \item How will the data be analysed?
    \begin{enumerate}[noitemsep,topsep=0pt]\small
        \item If automatic: what analysis tool will you use, and how will it be configured?
        \item If manual: what is the background of the annotators? How will you ensure reliability and validity of the analysis?
    \end{enumerate}
    \item What statistical tests will be used, if any?
    \item Anything else you'd like to preregister?
\end{enumerate}

\subsection{NLP Engineering experiment paper}\label{sec:questions-nlpexp}
NLP engineering experiments are like experiments in the social sciences, except that the subjects are NLP models and the performance data is model output. So the standard social science questions do not need to be modified that much to fit NLP experiments.

\begin{enumerate}[noitemsep,topsep=0pt]
\item What is the goal of your study?
\item What are your hypotheses/key assumptions?
\item What are the (in)dependent variables?
\item How will these variables be measured?
\item Is there just one condition, or more?
\item What software libraries will you use?
\item What hardware will you use?
\item What parameter settings will you use?
\item What data set will you use?
\item If the data set does not already exist, see \S\ref{sec:resource-prereg}. If it does:
\begin{enumerate}[noitemsep,topsep=0pt]\small
    \item How familiar are you with the data?
    \item To what extent are your hypotheses informed by yourself or others interacting with this data? To what extent does this hinder the generalisability of your approach?
    \item Are you planning to collect additional data to validate your approach?
\end{enumerate}
\item Why this data? What are its key properties?
\item How is the data split in train/val/test?
\item How will you analyse the results and test the hypotheses?
    \begin{enumerate}[noitemsep,topsep=0pt]\small
        \item If automatic: what metric(s) (including implementation) will you use, and how will they be configured?
        \item If human judgments: see \S\ref{sec:humaneval}.
    \end{enumerate}
\item Will you carry out an error analysis? If so, see~\S\ref{sec:erroranalysis}.
\item Anything else you'd like to preregister?
\end{enumerate}

\subsection{Position paper}
Position papers typically do not need to be preregistered, since they often do not provide any new data, but rely on the author's experience. These kinds of papers also usually signal that they are more opinionated than other kinds of papers.

\subsection{Reproduction paper}
For a reproduction paper, the questions are a mix of the questions above (\S\ref{sec:questions-nlpexp}) with reproduction-specific questions.

\begin{enumerate}[noitemsep,topsep=0pt]
\item What results do you aim to reproduce?
\item What kinds of experiments does this involve?
\item What is the goal?
\begin{enumerate}[noitemsep,topsep=0pt]\small
    \item What constitutes a successful reproduction?
    \item What constitutes an unsuccessful reproduction?
    \item What is the margin of error?
\end{enumerate}
\item Do you expect to be successful? Why (not)?
\item How are you planning to reproduce the original results?
\begin{enumerate}[noitemsep,topsep=0pt]\small
    \item Will you use the same soft/hardware?
    \item Will you use the same data?
    \item Will you use the same codebase?
    \item If human participants are used: will you target the same demographic, and use the same experimental settings?
    \item Will you contact the authors?
    \item How much time do authors have to respond to your queries?
    \item How much time/effort are you willing to spend?
\end{enumerate}
\item Will you carry out an error analysis? If so, see~\S\ref{sec:erroranalysis}.
\item Anything else you'd like to preregister?
\end{enumerate}

\subsection{Resource paper}\label{sec:resource-prereg}
It is at least a bit unexpected to promote preregistration for resource papers. After all, if all you do is data collection, then there are no hypotheses to test. But since the goal of this appendix is to provide a starting point for discussion, we are taking the stance that no study is free from biases or initial expectations. As such, it is useful to at least document what you aim to collect, for what reasons, and how you are planning to do so. Once the project is completed, if you 

\begin{enumerate}[noitemsep,topsep=0pt]
    \item What is the goal of this study?
    \item What kind of data will be collected?
    \item How will this data be collected?
    \item What is the intended application for the data you plan to collect?
    \item What sampling strategy will be used? Why?
    \item How much data are you planning to collect?
    (Is there any target or stopping criterion?)
    \item How will the data be analysed?
    \begin{enumerate}[noitemsep,topsep=0pt]\small
        \item If automatic: what analysis tool will you use, and how will it be configured?
        \item If manual: what is the background of the annotators? How will you ensure reliability and validity of the analysis?
    \end{enumerate}
    \item What properties should the data have?
    \item How will you ensure that the data will have those properties?
    \item Anything else you'd like to preregister?
\end{enumerate}

\subsection{Survey paper}
We would recommend that authors follow the PRISMA guidelines  \cite{10.1371/journal.pmed.1000097,10.1371/journal.pmed.1000100}) for their surveys. This requires authors to develop a review protocol, which means authors should answer the following questions before initiating their study:

\begin{enumerate}[noitemsep,topsep=0pt]
    \item What is the goal of this study?
    \item What is the rationale behind this study?
    \item What questions do you hope to answer?
    \item What types of articles are relevant to answer your question?
    \begin{enumerate}[noitemsep,topsep=0pt]\small
        \item What are the inclusion criteria?
        \item What are the exclusion criteria?
        \item What languages are included?
    \end{enumerate}
    \item How will you decide which articles are relevant? (e.g. judging by the title/abstract)
    \item What search engines will you use?
    \item What search queries will you use?
    \item What are the variables of interest?
    \item How will you synthesize the results?
    \item How will you ensure the reliability and validity of your study?
    \item Anything else you'd like to preregister?
\end{enumerate}

\subsection{Other kinds of preregistrations}
Instead of preregistering a full study, one might also preregister part of a study, e.g. a human evaluation or error analysis.

\subsubsection{Human evaluation}\label{sec:humaneval}
Human evaluation studies often do not report all the necessary details to reproduce their work \cite{howcroft-etal-2020-twenty}. Thus \citet{shimorina2021human} developed a datasheet for recording all the necessary details. This datasheet can mostly be filled in before the study is carried out. A selection of their questions is provided below (see the paper for more details and additional questions).

\begin{enumerate}[noitemsep,topsep=0pt]
    \item What type of input(s) does the system have?
    \item What type of output does the system produce?
    \item What task is the system supposed to carry out?
    \item What languages are involved?
    \item How many systems/outputs per system are being evaluated?
    \item How are the outputs selected?
    \item What is the statistical power of the sample size?
    \item What kind of evaluators are being used?
    \item What training is given to the evaluators?
    \item What is the background if the evaluators?
    \item How are responses collected?
    \item What quality assurance measures are used?
    \item What do evaluators see when carrying out evaluations?
    \item How free are evaluators regarding when and how quickly they are supposed to evaluate the results.
    \item Can evaluators provide feedback or not?
    \item What are the experimental conditions like?
    \item What type of quality is assessed in the evaluation?
    \item How is this quality assessed?
    \item How are the responses processed?
    \item What are the ethical implications of your work?
\end{enumerate}

\subsubsection{Error analysis}\label{sec:erroranalysis}
Error analysis is similar to human evaluation, except that it is typically more qualitatively oriented. This does not mean that there cannot be a quantitative component (e.g. counting the number of errors, comparing this number between different systems), but often systems are also just analysed by themselves, and we just want to know what future researchers still ought to improve about the system.

\begin{enumerate}[noitemsep,topsep=0pt]
\item What is the goal of the error analysis?
\item What type of input(s) does the system have?
\item What type of output does the system produce?
\item What task is the system supposed to carry out?
\item What languages are involved?
\item What do you know from the literature about system errors?
\item When does something count as an error?
\item What kinds of errors do you expect to find?
\item How many outputs will you analyse?
\item How will you sample the outputs to analyse?
\item Do you also consider the input in your sampling strategy?
\item How do you plan to analyse the output?
\item How many judges will assess the output? 
\item What is the background of the judges?
\item What training do the judges receive?
\item How is the reliability of the judges assessed?
\item How will their responses be processed?
\item Is there a fixed error categorisation scheme or not?
\end{enumerate}

\end{document}